\documentclass[11pt,pdfwrite]{article}
\usepackage{graphicx}
\title{Experiments with small helicopter automated landings at unusual attitudes}
\author{S. Bayrakar and E. Feron~\thanks{School of Aerospace Engineering, Georgia Institute of Technology, {\tt \{selcuk.bayraktar,feron\}@gatech.edu}}}
\begin{document}
\maketitle
{\bf Abstract:} This paper describes a set of experiments involving small helicopters landing automated landing at unusual attitudes. By leveraging the increased agility of small air vehicles, we show that it is possible to automatically land a small helicopter on surfaces pitched at angles up to 60 degrees. Such maneuvers require considerable agility from the vehicle and its avionics system, and they pose significant technical and safety challenges. Our work builds upon previous activities in human-inspired, high-agility flight for small rotorcraft. However, it was not possible to leverage manual flight test data to extract landing maneuvers due to stringent attitude and position control requirements. Availability of low-cost, local navigation systems requiring no on-board instrumentation has proven particularly important for these experiments to be successful.

\section*{Introduction}
Unmanned aerial vehicles have encountered growing popularity among civilian and military users alike, with hundreds of fixed-wing, unmanned aircraft operated every day throughout the world for surveillance and payload delivery missions~\cite{Aero:07}. With shrinking electronics size and increasing computer power, small airborne systems offer a wealth of new opportunities never imagined before the twenty-first century. The conjunction of growing interest for micro-air vehicles with their easy operation in small, laboratory-sized environments has led academia and industry to launch many research efforts aimed at improving their agility and explore the boundaries of their flight envelope. It may rightfully argued that high agility of small vehicles began with the era of missiles in the 1940's and 1950's. However, the novelty of the research opportunities offered by modern, small-sized machines is justified by the fact that (i) these machines are considerably under-actuated and (ii) many of them are expected to be recoverable, and to be able to land and spend significant amounts of time within the theater of operations before taking off again, much like birds do in nature. As a result, there is a strong incentive to study the possibility for small machines not only to fly well and navigate properly during flight, but also to land in possibly constrained locations for the purpose of either replenishing their resources or performing a surveillance task. Words such as ``perching'' have become part of the popular jargon associated with such research activities, and much effort (discussed thereafter) has been devoted to performing landing maneuvers in constrained environments. Leveraging the small size of the vehicles has led industry to come up with very imaginative solutions for the recovery of small vehicles. Among them, the ``skyhook'' concept developed by the Insitu Group (URL {\tt www.insitu.com/uas}) is a model of innovative landing system, now in use by the US Navy and Coast Guards. The system bypasses the need for costly and dangerous landing maneuvers on a large, moving deck, something that manned aircraft must perform. 
Despite the important operational relevance of helicopters, including their ability to hover and operate in cramped environments, 
limited progress has been made, however, towards improving their landing conditions. 

Among noted and recent research contributions to the helicopter landing problem we find results from a control systems perspective~\cite{KoS:98,SKHS:98,Isidori:03} and from a vision-based sensing perspective~\cite{SSS:01,SVSMS:02}. Other teams have integrated control and vision together to demonstrate landings in various conditions~\cite{Gaurav2:03,Gaurav3:03,Gaurav4:02,Corke1:04,TSR:05a,TVF:06a}. However, the available experimental research literature focuses on autonomous helicopters landing on horizontal or quasi-horizontal surfaces: From a control systems perspective, the achieved vehicle performance does not differ much from technology available as far back as the 1950's \\(see {\tt http://www.gyrodynehelicopters.com/} \\
or {\tt www.vtol.org/uavpaper/NavyUAV.htm} for examples). 
On the side of theory, probably one of the most significant recent research concerns the control of tethered helicopters during landing~\cite{Agrawal:05}, where the authors analyze the controlled dynamics of helicopters attached to a ship by means of a towing cable. 
On the side of operations, landing a helicopter on non-horizontal terrain is considered to be a difficult task. According to experienced helicopter pilots~\cite{root:07}, landing on sloped terrain requires lowering the helicopter nacelle down to the ground slowly, making sure that the rotor remains approximately horizontal as the nacelle slowly adjusts to local terrain orientation. This 
tends to produce significant structural stresses on the rotor of helicopters with semi-rigid or rigid hubs.

This paper focuses on the experimental demonstration of small helicopter landings at unusual attitudes (on surfaces inclined by as much as 60 degrees), and the control methods used to achieve this result. We believe such a contribution is a useful intermediate step towards enabling full, all-attitude vehicle landing or alighting in geometrically constrained areas. In particular, we believe 
our experimental work constitutes a useful step towards vehicle perching, much like birds and bats do.

This paper is organized as follows: First, we describe the experimental setup used to perform 
the research, including the research vehicle, available instrumentation, and airspace layout. Then we briefly report human-in-the-loop experiments for helicopter landing on sloped platforms. The principles of automated vehicle landing are then presented, together with the control laws that were designed. Finally, experimental results are presented. 

\section{Experimental setup}
The experimental setup used for helicopter landing consists of (a) the flight test article (b) the landing pad and (c) the instrumentation system.
\subsection{Flight test article}
The flight test article chosen for the experiment is the Robbe Eolo Pro electric helicopter. 
This machine can be purchased for a relatively low price, making it a good candidate 
for experimentation.
\begin{figure}[htbp]  
   \hspace{0mm}
   \begin{center}
    \includegraphics[width=5cm]{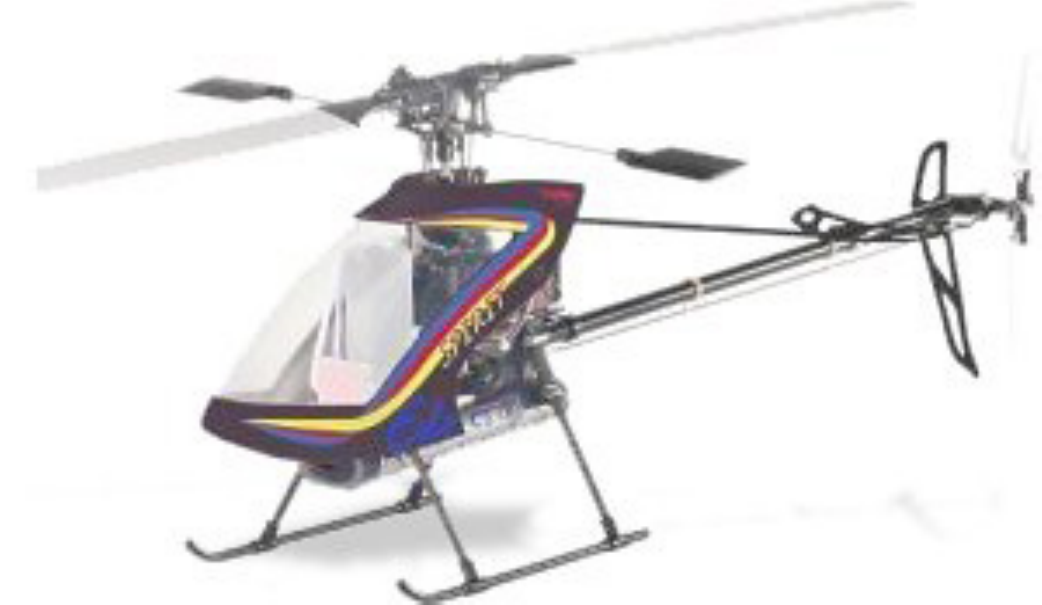}
    \includegraphics[width=7cm]{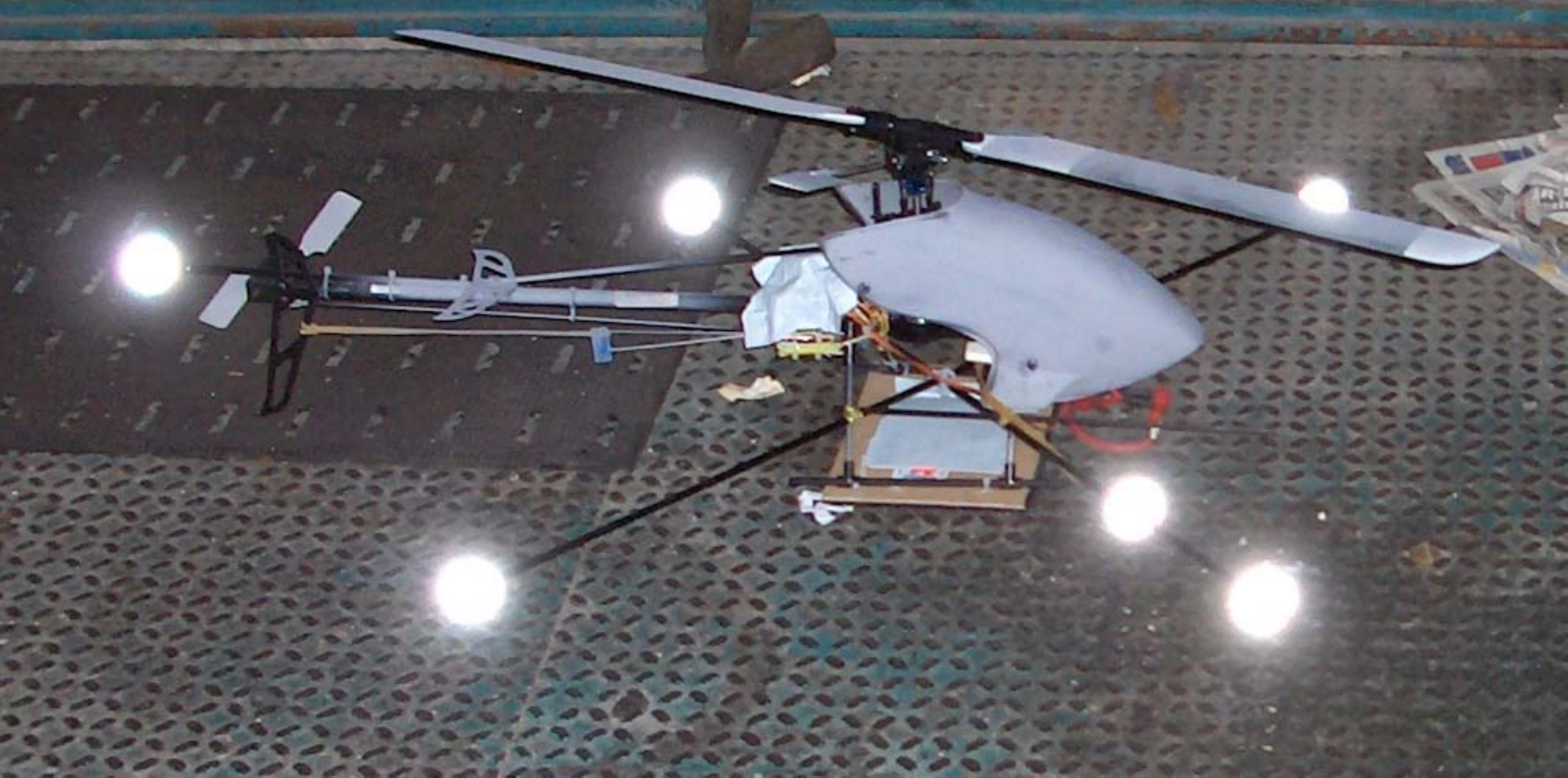}    
    \caption{Flight test article. Left: Commercial configuration. Right: Flight test configuration}
\label{helo}
\end{center}
\end{figure} 
The technical specifications of the machine used for our experiments are given in table below.

\begin{tabular}{|l|}\hline
Main rotor diameter: 870 mm\\ \hline
Tail rotor diameter: 178 mm\\ \hline
All-up weight: 1300 g \\ \hline
Height: 256 mm \\ \hline
Length: 725 mm  \\ \hline
Propulsion System: Brushless electric motor \\ 
Medusa Research MR-028-056-2800\\
(http://www.medusaproducts.com/motors)\\ \hline
\end{tabular}
\vspace{5mm}

It can also be seen from the pictures that the rotor features Bell-Hillier stabilizer bars. In addition, the engine RPM is held constant using an on-board governor.

\noindent For flight testing purposes, the helicopter has been modified as follows: 
First, the landing gear was modified to be compatible with the landing pad (described thereafter). 
Second, the helicopter was painted with matte, gray paint to avoid unwanted reflections that would have 
interfered with the ground-based navigation system. Moreover, lightweight carbon rods were added to the landing gear. These rods support highly reflective beacons (see Fig.~\ref{helo}, right), used by the navigation system.

\subsection{Instrumentation}
\begin{figure}[htbp]  
   \hspace{0mm}
   \begin{center}
    \includegraphics[width=6cm]{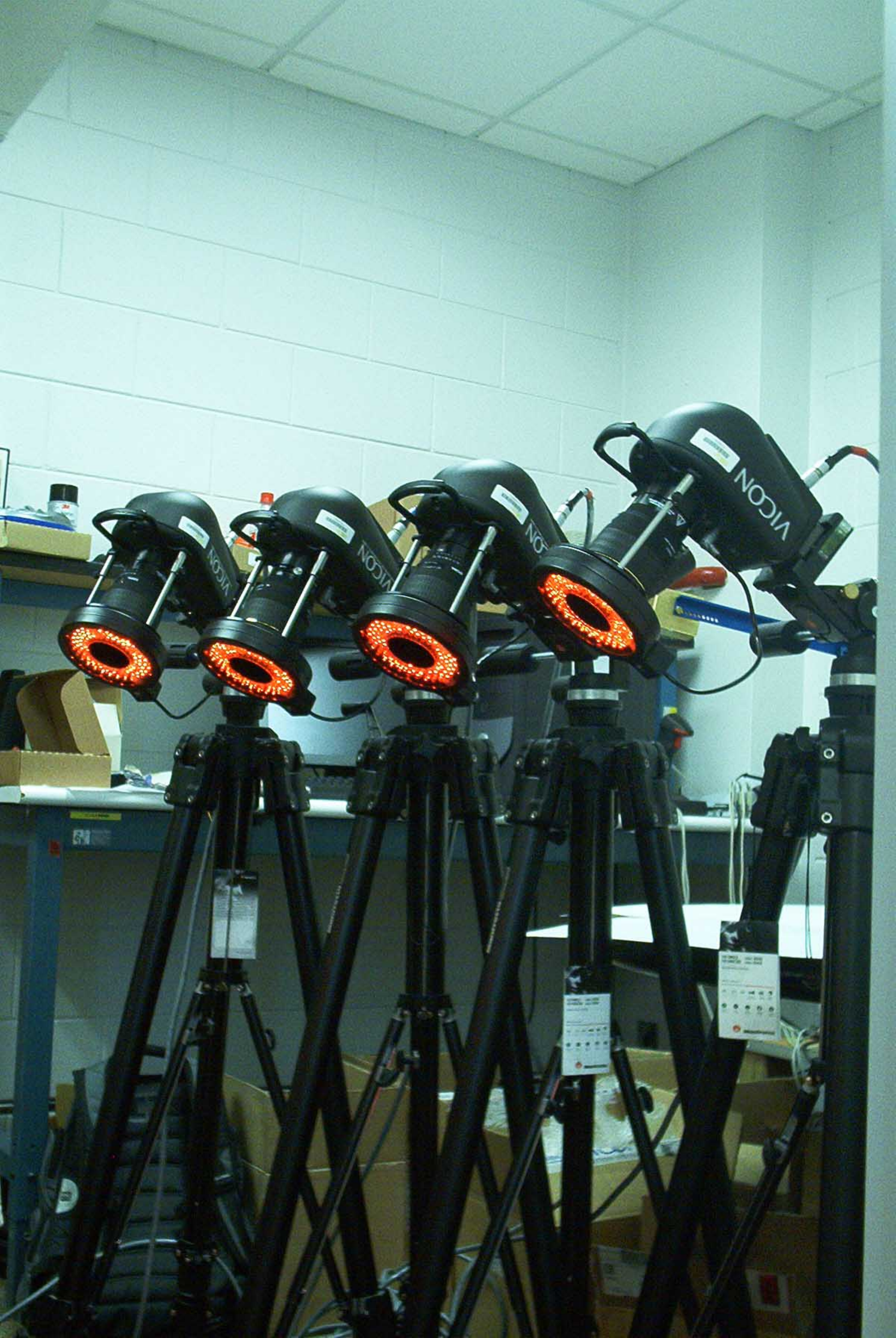}   
    \caption{Cameras of VICON positioning system}
\label{VICON}
\end{center}
\end{figure}

A key element that enables aggressive maneuvers during flight testing of small vehicles is the recent availability of reliable, ground-based navigation systems requiring only minimal on-board instrumentation. The VICON system provide such a turnkey navigation system. The system was originally designed to track human and animal motion.
Its ground infrastructure consists of several cameras (typically six or more) which actively illuminate the object to be tracked via LEDs (see Fig.~\ref{VICON}). With highly reflective coating over designated markers such as small-sized, styrofoam balls, the VICON system can track one or several rigid bodies in terms of position and orientation~\cite{VBFHF:06}. 
For the purpose of avoiding any geometric ambiguity, it is important to place markers on the vehicle so as to break any symmetry (see Fig.~\ref{helo}). With such precautions taken and with enough cameras, vehicle position and orientation can be obtained with centimeter position accuracy and sub-degree attitude accuracy, following a short calibration procedure. 

\subsection{Landing pad}
\begin{figure}[htbp]  
   \hspace{0mm}
   \begin{center}  
     \includegraphics[width=8cm]{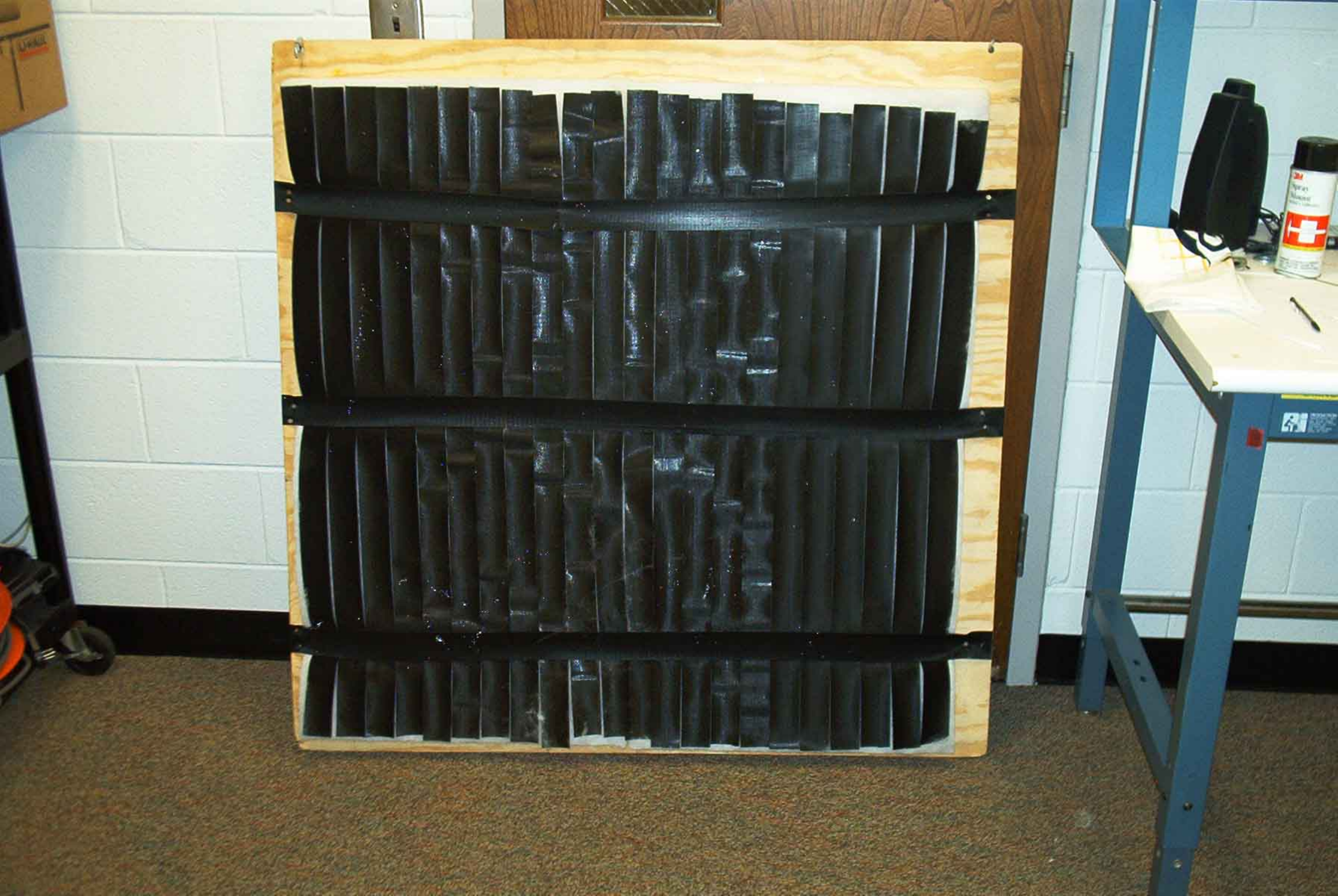}
    \caption{Landing pad}
\label{landing_pad}
\end{center}
\end{figure}

The landing pad is shown in Fig.~\ref{landing_pad} and consists of a square, 1.2m $\times$ 1.2m piece of plywood covered with Velcro. This piece of plywood may be easily tilted at various angles.

Complementary Velcro material is mounted under the helicopter landing skids, so that upon contact the two elements (the landing pad and the helicopter landing gear) would effectively bond. While it may be argued that such a solution creates artificially favourable conditions for successful landings, we consider that it constitutes a useful intermediate step toward performing landings in unequipped areas at unusual attitudes. In addition, with the development of Velcro-like 
material made out of shape-memory alloys~\cite{eads:07}, this option may even prove viable in several operational instances.

\subsection*{Experimental layout}
\begin{figure}[htbp]  
   \begin{center}  \hspace{-25mm}
      \includegraphics[width=15cm]{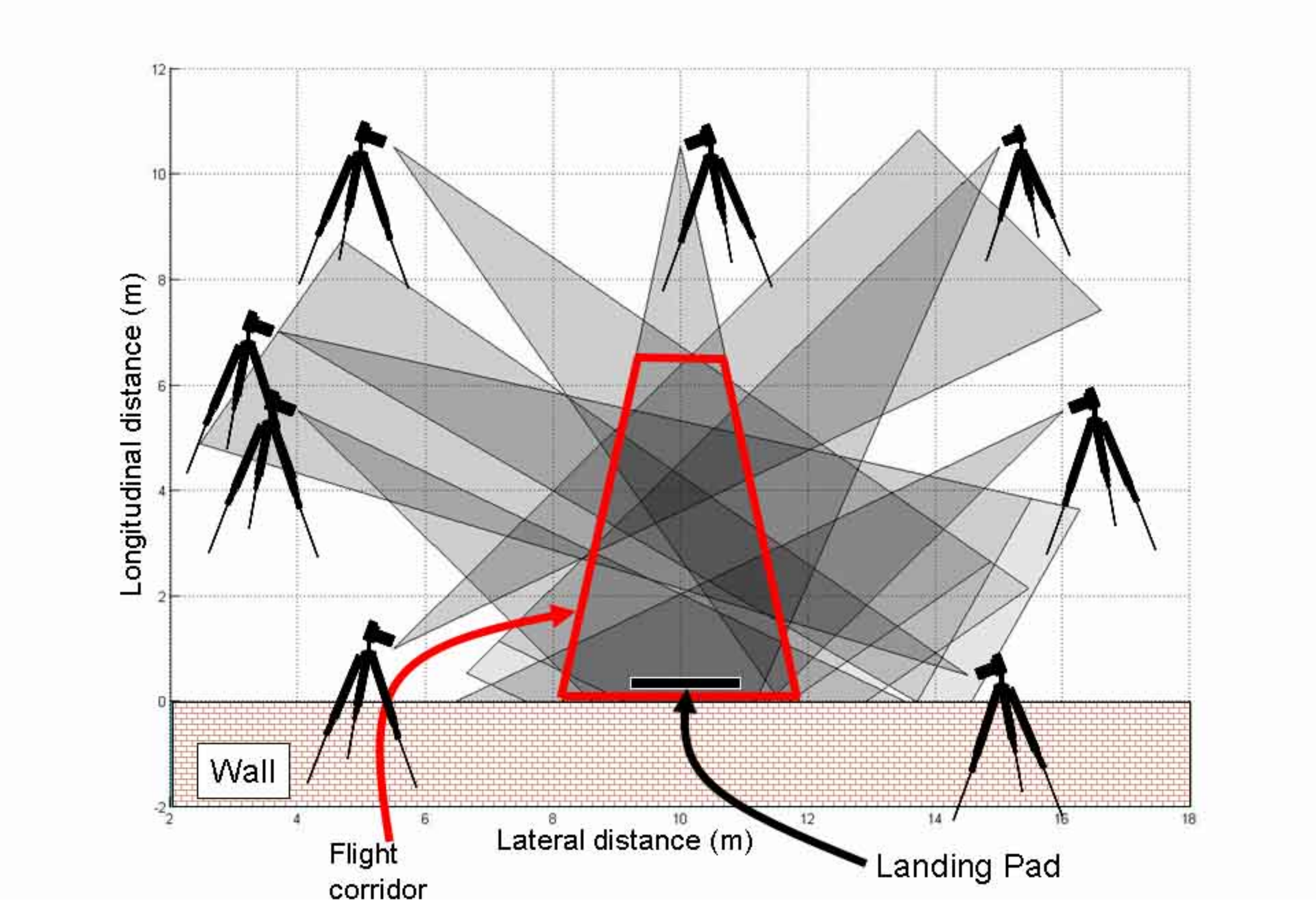}   
    \caption{Camera Layout}
\label{cam_layout}
\end{center}
\end{figure}

Flight tests were conducted in the courtyard of Georgia Tech's School of Aerospace, a closed area used primarily as temporary parking space. A top view of the camera layout is given in Fig.~\ref{cam_layout}.
The VICON cameras were located in such a way that accurate position, velocity, attitude and angular velocity information could be obtained for the helicopter in a corridor
containing the helicopter initial position and the landing area.
This corridor is 6 meters deep, 1 meter wide at maneuver inception and greater than 3 meters wide near the landing area. The available corridor height is about 3 meters.

\section{Experiments with humans in the loop}
The first set of experiments involved an expert human pilot aiming at landing the helicopter on a moderately pitched target (25 degrees). The human pilot was chosen for his ability to perform advanced aerobatic maneuvers (including inverted flight, loops, flips, rolls etc.). The flight was performed at night, which challenged the pilot's ability to precisely locate the helicopter relative to the target. In particular, the strategy followed by the pilot to land the helicopter failed to exhibit many of the desired characteristics for landing at high pitch angles: The precise navigation requirements associated with helicopter landing forced the pilot to give up on adjusting the helicopter pitch angle so as to match that of the landing pad. Instead, the pilot chose to hover the helicopter horizontally above the landing pad, and then dropped the helicopter on the pad by bringing the collective control down. This experiment showed that previously developed human-inspired strategies for aggressive flight control~\cite{GFMPF:02,GMF:04,Abb:07a,Ng:04a} could not be applied to the task described in this paper.

One of the benefits of the piloted experiments, however, was to demonstrate the validity of the ``Velcro'' landing pad concept, since the helicopter successfully and systematically bonded with the landing pad.

\section{Landing maneuver design}
This section presents our landing maneuver design philosophy, followed by a detailed presentation of the control laws used to perform the maneuver.

\subsection{Overall philosophy}
\begin{figure}[htbp]  
   \hspace{0mm}
   \begin{center}
    \includegraphics[width=12cm]{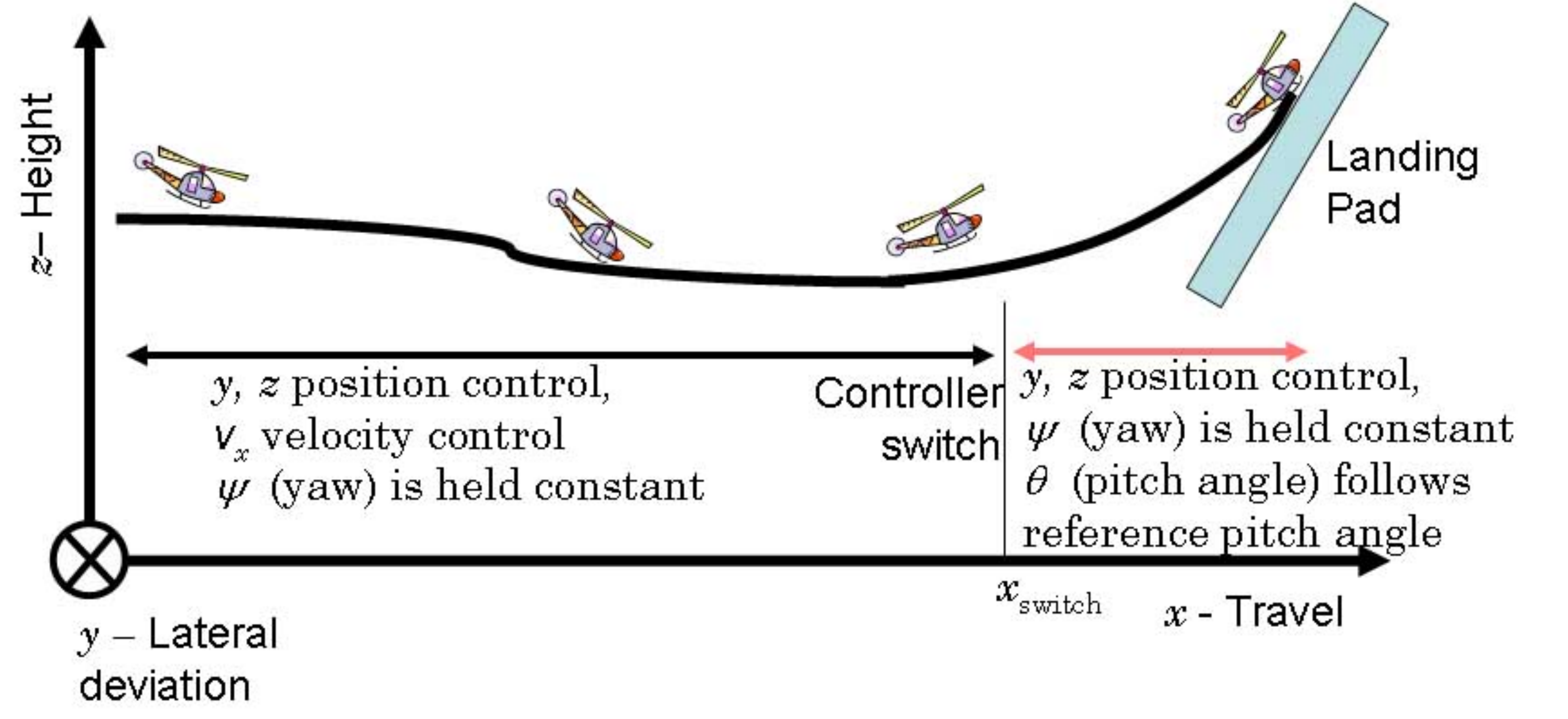}
    \caption{Landing procedure}
\label{landing_procedure}
\end{center}
\end{figure}

The landing maneuver was designed as a two phase process, shown in Fig.~\ref{landing_procedure}.
What makes the landing maneuver challenging is the underactuated nature of the helicopter. Assuming constant helicopter rotor speed, four actuation mechanisms are available (collective, pitch and roll cyclic, and yaw). However, six degrees of freedom must be controlled (geometric position + attitude). Our landing maneuver is divided in two phases, whereby

\begin{itemize}
\item During approach, the lateral position $y$, altitude $z$, forward velocity $\dot{x}$ and yaw angle $\psi$ of the helicopter are primarily controlled. 
\item During flare and eventual landing, the pitch ($\theta$) and yaw ($\psi$) angles of the helicopter are primarily controlled, together with the helicopter lateral position $y$ and altitude $z$. 
\end{itemize}

Thus, the vehicle undergoes a controller mode switch during the maneuver so as to enable the landing maneuver. This mode switch is reminiscent of strategies previously used in previous demonstrations of aerobatic flight by miniature helicopters~\cite{GFMPF:02,vlad:two}. The mode switch is triggered when the longitudinal vehicle position $x$ crosses a specific threshold $x_{\rm switch}$.

\subsection{Fail safe landing procedure}
Owing to sensing limitations and the inherent risks associated with landing at high pitch angle, a fail-safe landing sequence was designed so that the vehicle could recover in case of a missed landing: Following an approach similar to that used in fixed-wing Naval operations, an abort maneuver was programmed to be executed whether the landing actually occurs or not. This procedure was based on the observation that once the velcro pads on the landing pad and on the helicopter are in contact, they provide such a strong bond that any follow-on control sequence from the abort procedure applied to the helicopter is ineffective at inducing any significant helicopter motion. On the other hand, should the helicopter fail to get in contact with the landing pad, the abort procedure can fully recover the helicopter and prevent a crash: Indeed, at high pitch angles, missed contact with the target means that the helicopter immediately enters a glide motion down towards the ground, that must be handled immediately.  In that regard, the procedure is similar to some fixed-wing aircraft naval aircraft operations, whereby landings are followed by the beginning of an aborted landing procedure in case the aircraft tail hook fails to catch one of the transverse cables.

\subsection{Control architecture}

\begin{figure}[htbp]  
   \hspace{0mm}
   \begin{center}
    \includegraphics[width=12cm]{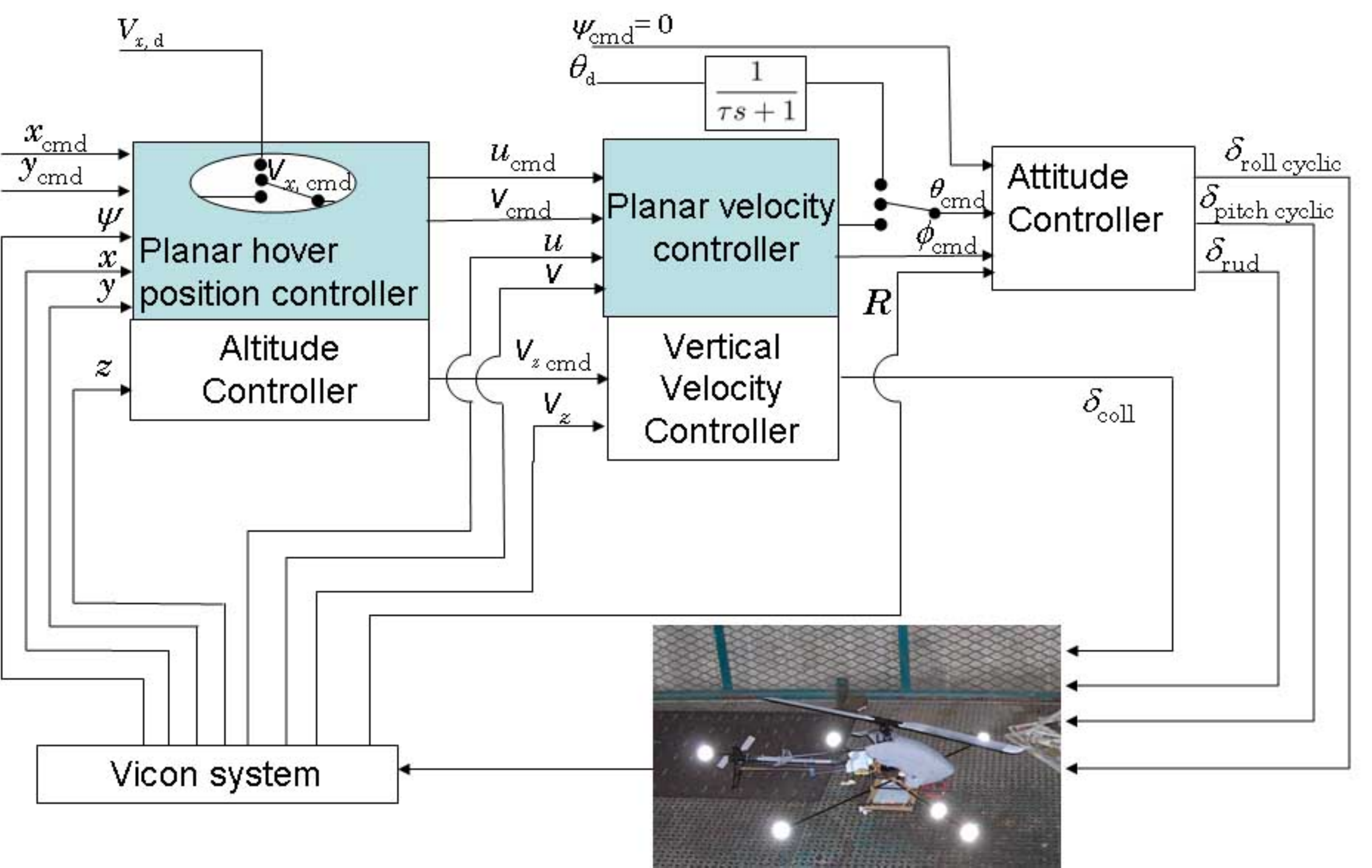}
    \caption{Helicopter control architecture}
\label{architecture}
\end{center}
\end{figure}

The control architecture is given by the diagram in Fig~\ref{architecture}. Depending on the phase of the flight,
only part or all of the controller architecture is used: For example, when hover control is desired, the entire control architecture is used. However, the vehicle forward velocity $v_x$ may be controlled directly as well. Likewise, the helicopter pitch angle can be controlled directly. 

Three controllers are mounted in series, corresponding to several successive loop closures. First, the {\em position controller} controls the forward and lateral motions of the helicopter, corresponding to the horizontal absolute coordinates $x$ and $y$. The control laws (in terms of commanded speeds $v_{x,\; \rm cmd}$, $v_{y,\; \rm cmd}$, $v_{z,\;\rm cmd})$ are proportional laws with saturation:
\[
\begin{array}{rcl}
v_{x,\rm cmd} &=& -{V_{\rm max}} \mbox{SAT}\displaystyle \left(\frac{\lambda_p (x-x_{\rm cmd})}{V_{\rm max}}\right)\\[10pt]
v_{y,\rm cmd} & = & -{V_{\rm max}} \mbox{SAT}\displaystyle \left(\frac{\lambda_p (y-y_{\rm cmd})}{V_{\rm max}}\right)\\[10pt]
v_{z,\rm cmd} & = & -{V_{\rm max}} \mbox{SAT}\displaystyle \left(\frac{\lambda_z (z-z_{\rm cmd})}{V_{\rm max}}\right)
\end{array}
\]
where $x_{\rm cmd}$, $y_{\rm cmd}$, and $z_{rm cmd}$ are the commanded positions and $x$, $y$, $z$ are the position of the helicopter in inertial coordinates. The operators SAT are the usual saturation function. Alternatively, the $x$-axis velocity $v_x$ may be controlled directly as shown in Fig.~\ref{architecture}.
The commanded velocities are then converted from an inertial reference frame to the helicopter reference frame
\[\begin{array}{rcl}
\left[\begin{array}{c} u_{\rm cmd} \\ v_{\rm cmd}\end{array} \right] 
&=& \left[\begin{array}{rc}\cos \psi & \sin \psi \\-\sin \psi & \cos \psi\end{array}\right]\left[\begin{array}{c} v_{x,\rm cmd}\\v_{y,\rm cmd}
\end{array}\right]
\end{array}\]
In these expressions, $\lambda_p$ and $\lambda_z$ are appropriately chosen positive constants.

The planar velocity command loop is computed as follows. A second-order velocity response is sought,
such that 
\begin{equation}
\begin{array}{rcl}
\displaystyle \frac{d}{dt}{u} & = & \lambda_u (u_{\rm cmd}-u+i_u )\\[10pt]
\displaystyle \frac{d}{dt}{v} &= &\lambda_v (v_{\rm cmd}-v+i_v)\\[10pt]
\displaystyle \frac{d}{dt}{i}_u & = & k_{i_u}(u_{\rm cmd}-u)\\[10pt]
\displaystyle \frac{d}{dt}{i}_v & = & k_{i_v}(v_{\rm cmd}-v).
\end{array}
\label{des_dyn}
\end{equation}
The presence of integrators (with states $i_u$ and $i_v$) aims at guaranteeing zero tracking error. 
In these expressions, $\lambda_u$, $\lambda_v$, $k_{i_u}$ and $k_{i_v}$ are appropriately chosen positive constants.
Assuming the simplified helicopter planar dynamics
\begin{equation}
\begin{array}{rcl}
\displaystyle \frac{d}{dt}{u} &=& \frac{1}{m}(-T_{\rm col} \theta_{\rm cmd}+k_{\rm du}\left|u\right|u)\\[10pt]
\displaystyle \frac{d}{dt}{v} &=& \frac{1}{m}(T_{\rm col} \phi_{\rm cmd} + k_{\rm dv} \left|v \right|v),
\end{array}
\label{dynamics}
\end{equation}
where the mass of the helicopter $m$ and the aerodynamic constants $k_{\rm du}$ and $k_{\rm dv}$ are determined experimentally,
and inverting the dynamics leads to expressions for $\theta_{\rm cmd}$ and $\phi_{\rm cmd}$
\[
\begin{array}{rcl}
\theta_{\rm cmd} &=& -\frac{1}{T_{\rm col}} (\lambda_u m (i_u + u_{\rm cmd}-u)- k_{\rm du} u \left|u\right|)\\
\phi_{\rm cmd} &= & \frac{1}{T_{\rm col}} (\lambda_v m (i_v+v_{\rm cmd}-v)-k_{\rm dv} v \left|v\right|)
\end{array}
\]
by combining~(\ref{des_dyn}) and~(\ref{dynamics}) together. Alternatively, the pitch angle $\theta$ may be directly commanded via the input $\theta_{\rm d}$. To avoid high-frequency excitation of the helicopter pitch cyclic, $\theta_{\rm d}$ is subject to low-pass (first-order) filtering.
Together with the desired heading $\psi_{\rm cmd}$, the Euler angles $\theta_{\rm cmd}$ and $\phi_{\rm cmd}$ allow us to form the desired attitude, expressed by the rotation matrix $R_{\rm cmd}$ (the reader is invited to consult
standard textbooks~\cite{Nel:98} for the expression of the rotation matrix $R_{\rm cmd}$ as a function of $\phi_{\rm cmd}$, $\theta_{\rm cmd}$, and $\psi_{\rm cmd}$). Computing the rotation error mnatrix $R_{\rm err} = R_{\rm cmd}^TR$ and computing $\epsilon_{\rm err}= \log R_{\rm err}$ (here the $\log$ operation applies to matrices as the inverse of the matrix exponential), we obtain
\[\epsilon_{\rm err} = 
\left[\begin{array}{ccc} 0 & -\epsilon_z & \epsilon_y \\
\epsilon_z & 0 & -\epsilon_x \\
-\epsilon_y & \epsilon_{x} & 0 \end{array}
\right].
\]
Standard results about rotations indicate that the orientation error $R_{\rm err}$ is a rotation whose axis is $[\epsilon_x \;\; \epsilon_y \;\; \epsilon_z]^T$ and whose angular amplitude is $\sqrt{\epsilon_x^2+\epsilon_y^2+\epsilon_z^2}$. The proportional 
control law
\[
\begin{array}{rcl} \delta_{\rm roll cyclic} = k_{\phi} \epsilon_x\\
\delta_{\rm pitch cyclic} = k_{\theta} \epsilon_y\\
\delta_{\rm rud} = k_{\rm \psi} \epsilon_z\end{array}
\]
was chosen for its simplicity and ability to steer $R$ towards $R_{\rm cmd}$. Damping is provided by the helicopter dynamics, including the Hillier stabilizer bars and the built-in gyroscopic yaw damper.

\begin{figure}[htbp]  
   \hspace{0mm}
   \begin{center}
    \includegraphics[width=12cm]{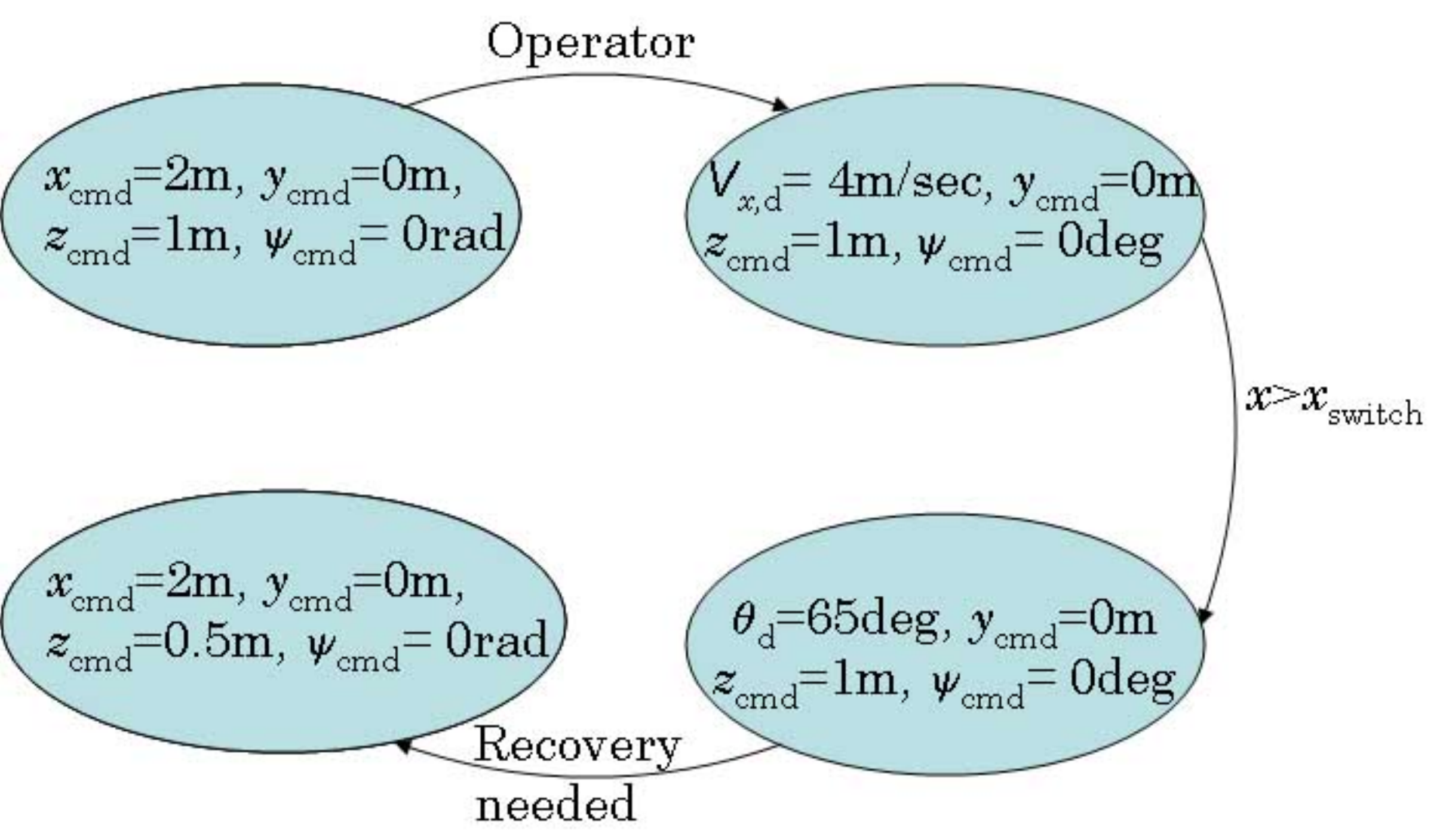}
    \caption{Landing maneuver sequence}
\label{automaton}
\end{center}
\end{figure}

The velocity controller about the $z$ inertial axis was computed 
using the desired behavior
\[
\begin{array}{rcl}
\displaystyle \frac{d}{dt}{v}_z &=& \lambda_{v_z} (v_{z,\rm cmd}-v_z+ i_{v_z})\\[10pt]
\displaystyle \frac{d}{dt} {i}_{v_z} & = & k_{i_{v_z}}(v_{z, \rm cmd}-v_{z}).
\end{array}
\]
Using the simplified vertical dynamics
\[
\dot{v}_z = \frac{1}{m}(-T_{\rm coll} \cos \theta \cos \phi + mg - k_{\dot{z}}v_z),
\]
and
\[
T_{\rm coll} = k_{\rm coll} \delta_{\rm coll}
\]
yields the desired control law
\[
\delta_{\rm coll} = \frac{gm-\lambda v_z m (i_{v_z}+v_z-v_{z, \rm cmd})-k_{\dot{z}}v_z}
{k_{\rm coll}\cos \theta \cos\phi}.
\]
Such a control architecture enables a high-level, discrete control of the helicopter: The landing sequence may be seen as a discrete sequence of step inputs to the control architecture as shown in Fig.~\ref{automaton}.

\section*{Experimental results}
\begin{figure}[htbp]  
   \hspace{0mm}
   \begin{center}
    \includegraphics[width=12cm]{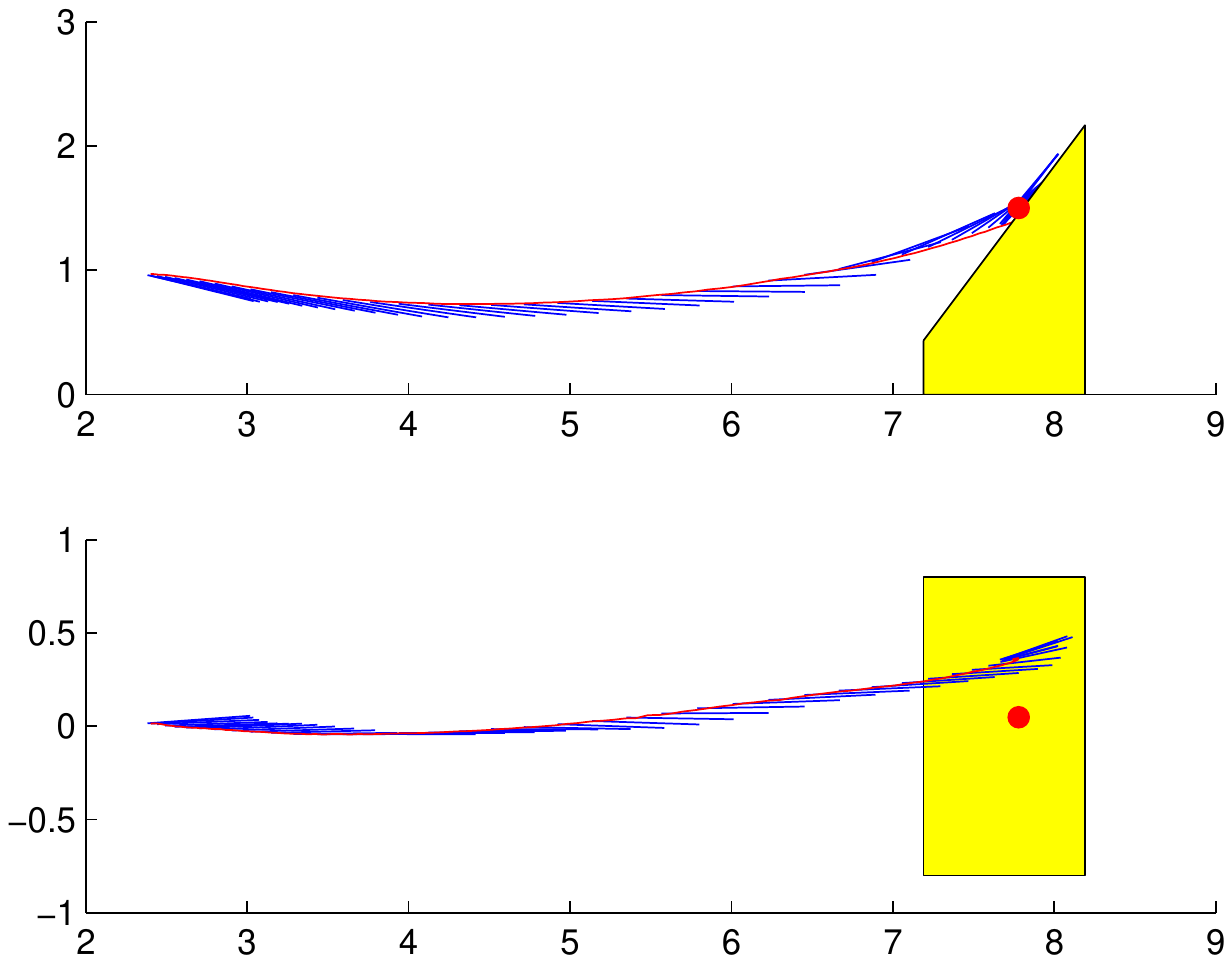}
    \caption{Successful landing: Top: Side view. Bottom: Top view. Helicopter attitude is represented by blue segments. Consecutive segments are separated by 0.06 sec.}
\label{touchdown}
\end{center}
\end{figure}

\begin{figure}[htbp]  
   \hspace{0mm}
   \begin{center}
    \includegraphics[width=12cm]{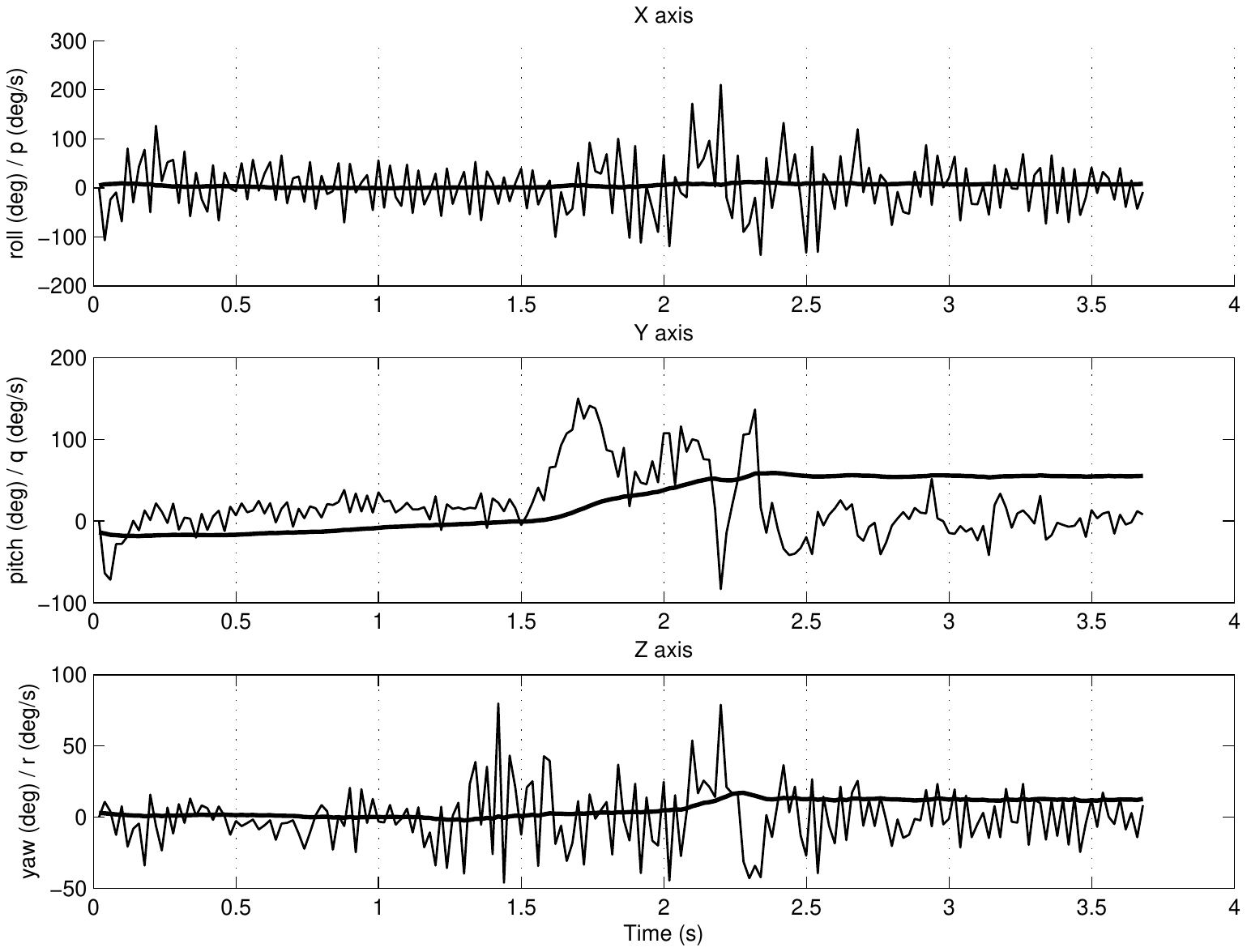}
    \caption{Successful landing: Helicopter attitude/attitude rates.}
\label{attitude}
\end{center}
\end{figure}

\begin{figure}[htbp]  
   \hspace{0mm}
   \begin{center}
    \includegraphics[width=12cm]{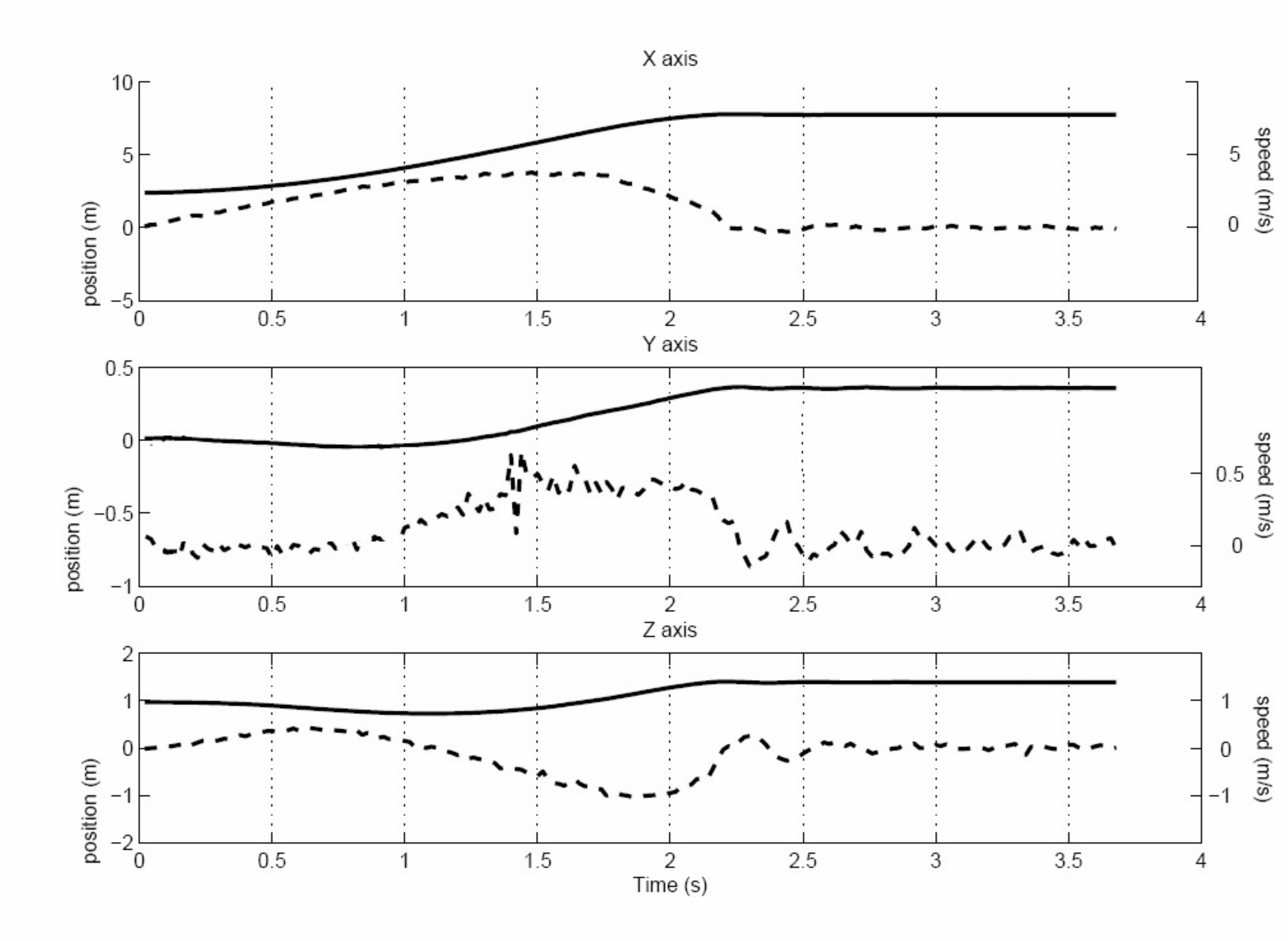}
    \caption{Successful landing: Helicopter position/velocity. Top: Along track position/velocity; Middle: Cross-track position/velocity. Bottom: Altitude / vertical speed.}
\label{position}
\end{center}
\end{figure}

\begin{figure}[htbp]  
   \hspace{0mm}
   \begin{center}
    \includegraphics[width=12cm]{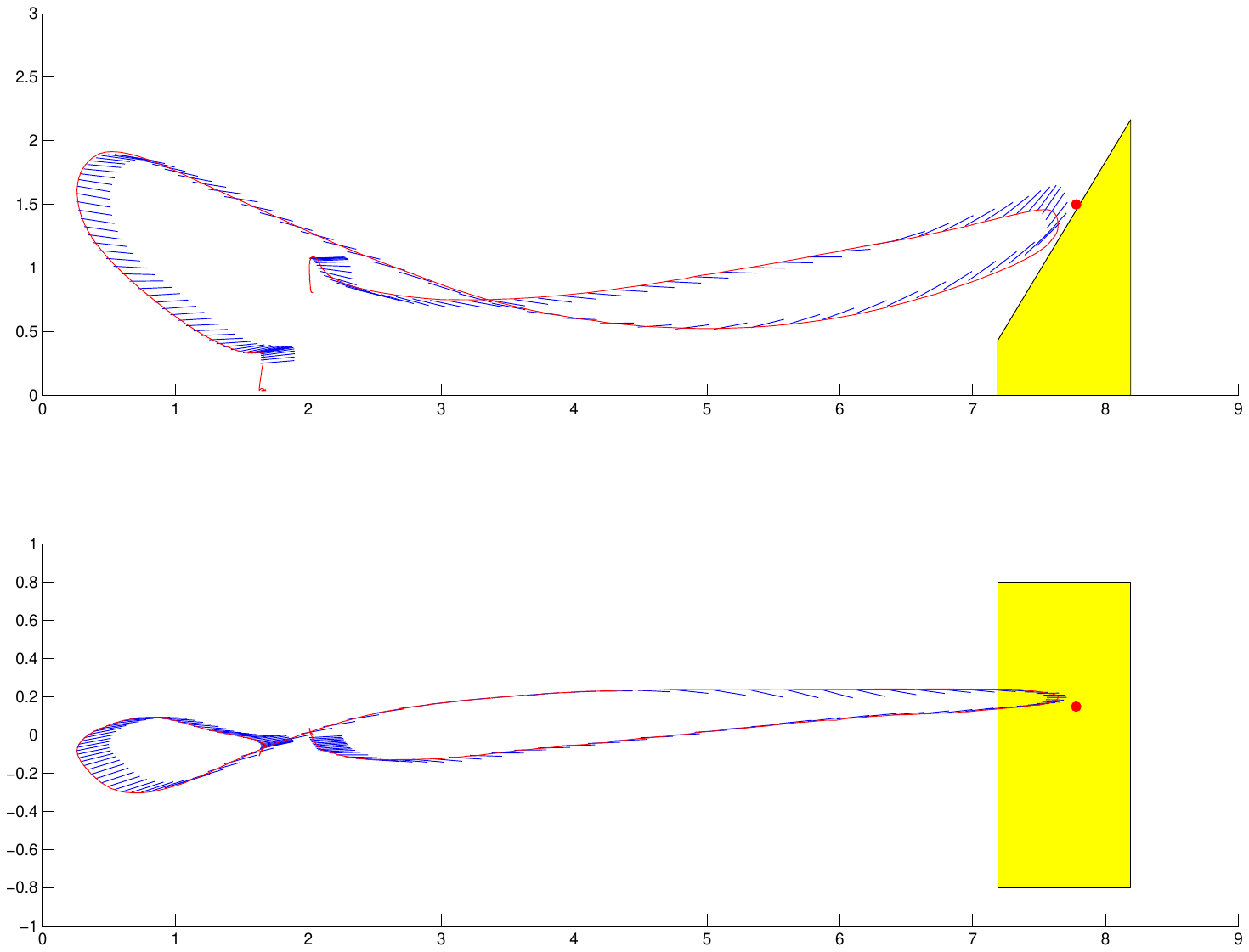}
    \caption{Aborted landing: Top: Side view. Bottom: Top view. Helicopter attitude is represented by blue segments. Consecutive segments are separated by 0.06 sec}
\label{aborted}
\end{center}
\end{figure}

The controlled helicopter was able to perform several landings, eventually reaching 60 degrees pitch angle at landing. Initial landings were performed at much smaller pitch angles (eg 10 degrees) for the purpose of calibrating the landing procedure and the abort process, should it be needed. The pitch angle of the landing pad was then progressively increased so as to eventually reach 60 degrees. Fig.~\ref{touchdown} shows the trajectory followed by the helicopter during one a successful landing.
The corresponding time histories for attitude/attitude rates and position/speeds are given in Fig.~\ref{attitude} and Fig.~\ref{position}.
It must be noted that the landing maneuver is, so far, performed ``open-loop'', that is the exact location of the landing pad was not measured in real-time, leading to several aborted maneuvers. Fig.~\ref{aborted} illustrates one such aborted maneuver.
Starting from a hovering position of at the location $(2,1)$ meters, the vehicle misses the target and executes an abort maneuver. The resulting maneuver is fairly ample and the vehicle violates the boundaries of the flight corridor. 

\section{Conclusion and further research}

This note presented the first published landing of a small helicopter at high pitch angle, 
with landings on platforms with up to 60 degrees pitch. Such maneuvers, together with other aggressive maneuvers, contribute to forming the core knowledge that is necessary to enable bird-like behaviors for small unmanned air vehicles. Such behaviors include the ability for these vehicles to alight at unusual attitudes. This interim result shows that these high-agility behaviors are within reach of available vehicles and navigation systems. Further research will include the ``robustification'' of the landing maneuvers by effectively providing the vehicle with on-line feedback of its position relative to the target. It will also include reaching higher pitch angles, including partially inverted landings on overhanging landing sites and landings on moving platforms. However, such activities will require a more comprehensive positioning system since such maneuvers exceed our current navigation capabilities.

\section*{Acknowledgements}
This research was supported by the Office of Naval Research under Grant N00014-06-1-1158. The authors would like to thank H. DeBlauwe, M. Gariel, T. Hunter, F. Lokumcu, V. Stojanovska, R. Valenzuela, and the safety pilot J. Fine for their support during this effort.

\newcommand{\etalchar}[1]{$^{#1}$}

\end{document}